\documentclass[letterpaper, 10 pt, conference]{ieeeconf}
\IEEEoverridecommandlockouts
\usepackage{graphicx}
\usepackage{amsmath}
\usepackage{amssymb}
\usepackage{booktabs}
\usepackage{caption}
\usepackage{subcaption}
\usepackage{enumitem}
\usepackage{multicol}
\usepackage{multirow}
\usepackage[pagebackref,breaklinks,colorlinks]{hyperref}
\usepackage[capitalize]{cleveref}
\usepackage[square,numbers,compress]{natbib}
\usepackage{amssymb}
\usepackage{pifont}
\usepackage[normalem]{ulem}
\useunder{\uline}{\ul}{}

\crefname{section}{Sec.}{Secs.}
\Crefname{section}{Section}{Sections}
\Crefname{table}{Table}{Tables}
\crefname{table}{Tab.}{Tabs.}

\begin{document}

\title{Out-of-Distribution Detection for LiDAR-based 3D Object Detection}

\author{Chengjie Huang $\dagger$, Van Duong Nguyen $\dagger$,
\thanks{$\dagger$ These two authors contribute equally to the work.}
Vahdat Abdelzad, Christopher Gus Mannes,\\ 
Luke Rowe, Benjamin Therien, Rick Salay, and Krzysztof Czarnecki\\
}

\maketitle
\begin{abstract}

3D object detection is an essential part of automated driving, and deep neural
networks (DNNs) have achieved state-of-the-art performance for this task.
However, deep models are notorious for assigning high confidence scores to
out-of-distribution (OOD) inputs, that is, inputs that are not drawn from the
training distribution. Detecting OOD inputs is challenging and essential for the
safe deployment of models. OOD detection has been studied extensively for the
classification task, but it has not received enough attention for the object
detection task, specifically LiDAR-based 3D object detection. In this paper, we
focus on the detection of OOD inputs for LiDAR-based 3D object detection. We
formulate what OOD inputs mean for object detection and propose to adapt several
OOD detection methods for object detection. We accomplish this by our proposed
feature extraction method.
To evaluate OOD detection methods, we develop a simple but effective technique of
generating OOD objects for a given object detection model. Our evaluation based
on the KITTI dataset shows that different OOD detection methods have biases toward
detecting specific OOD objects. It emphasizes the importance of combined OOD
detection methods and more research in this direction.
\end{abstract}
\section{Introduction}
\label{sec:intro}

\begin{figure}
    \centering
    \includegraphics[width=0.7\linewidth]{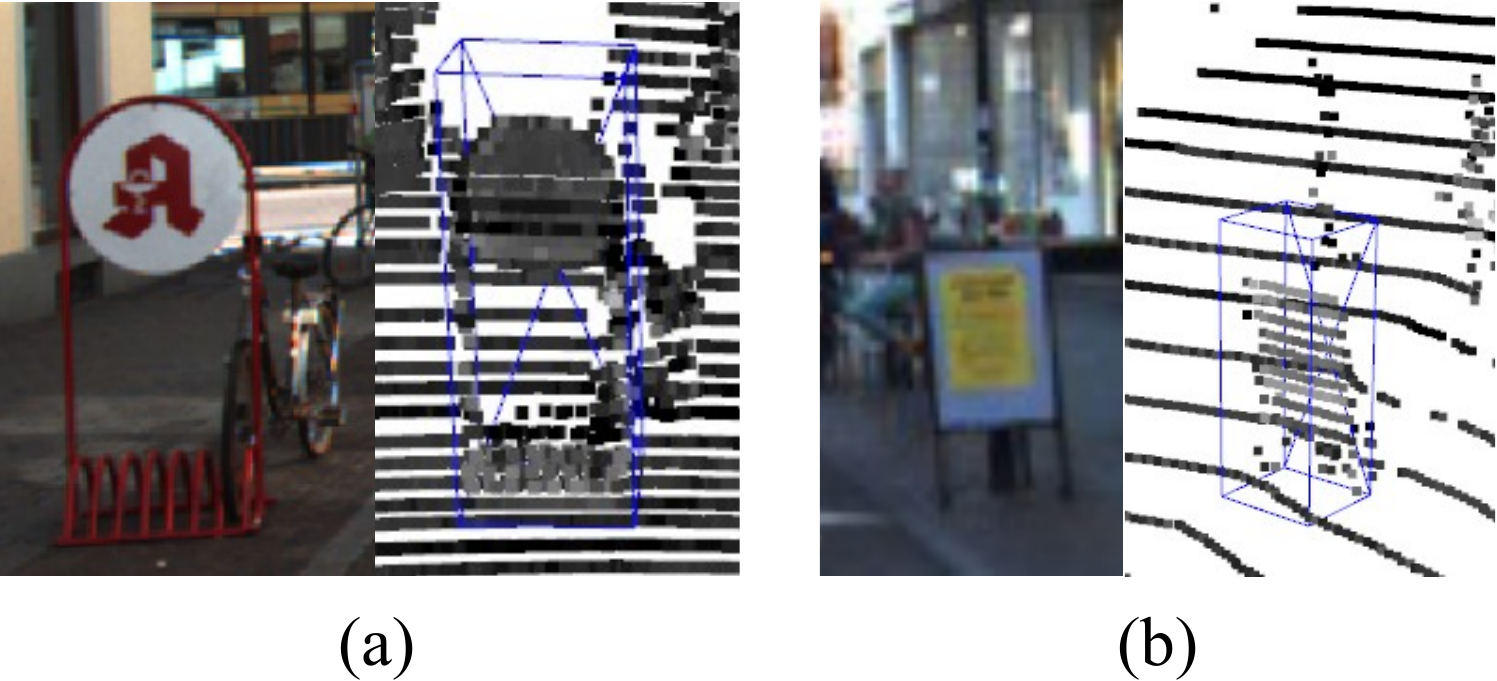}
    \caption{Examples of OOD objects detected as foreground objects by
    PointPillars \cite{lang2019pointpillars}. The camera images are for
    visual reference only; the corresponding fragments of the input point clouds
    are to the right. (a) A bike rack is detected as a pedestrian with 0.86
    confidence. (b) A signboard is detected as a pedestrian with 0.74
    confidence.}
    \label{fig:motivation}
    \vspace{-0.2in}
\end{figure}

3D object detection is a crucial part of autonomous vehicles (AVs).
However, state-of-the-art (SOTA) LiDAR-based 3D object
detection methods rely on deep neural networks~\cite{zheng20213dsessd,shi2020pvrcnn,lang2019pointpillars},
which are susceptible to the out-of-distribution (OOD) problem: they can make predictions with
high confidence given an input not drawn from the training distribution (not ID). It poses a safety concern that can hinder the deployment of AVs on public roads.
For instance, misdetection of a bike rack or signboard on the roadside as a
pedestrian (as shown in \Cref{fig:motivation}) can cause the AV to apply a hard brake or other
potentially dangerous evasive maneuvers. OOD detection
\cite{salehi2021unified-review-ood} aims to detect such cases.

OOD detection in the context of 3D object detection has hardly
been explored.
Existing OOD detection research focuses mainly on image classification
\cite{salehi2021unified-review-ood,liang2017odin-cls,
hendrycks2019-selfsupervised,rezende2015flow-cls,lee2018mahalanobis-cls,
hendrycks2016baseline-cls,ren2019likelihoodood-generative-cls,
zhang2020hybrid-cls,lee2017confidence-gan-cls,vyas2018ood-uncertainty-cls,
laksh2016deepensemble-uncertainty-cls,li2017dropout-uncertainty-cls,
hsu2020gen-odin-cls} or segmentation tasks
\cite{bevandic2018segmentation,angus2019segmentation,marchal2020segmentation,
williams2021segmentation}.
However, there are challenges unique to the object detection task
that such methods do not address.

The definition of OOD in classification cannot be directly applied to
object detection. In classification, the training distribution consists of a 
finite set of classes, and samples belonging to other classes are considered as OOD. For example, for the MNIST handwritten digit dataset
\cite{lecun-mnisthandwrittendigit-2010}, any image that does not contain 
handwritten digit is OOD. In object detection, however, in addition to a
finite number of foreground (FG) classes, the model is also
trained on a highly heterogeneous background (BG) class
which includes all objects that do not belong to the FG classes.
Since the union of FG and BG classes includes all possible objects, 
using the same OOD definition from classification would imply
that all objects are ID and no object is OOD.


Furthermore, in contrast to image classification, inputs to object detection may
contain multiple objects. It brings an extra challenge for OOD detection methods that rely on raw inputs
or feature maps.
While OOD detectors for classification can use the
entire input or feature map, object detectors may produce an
arbitrary number of
predictions per input, and thus the OOD detectors need to extract
inputs or features
associated with each prediction.

Lastly, evaluating OOD detection methods requires access to OOD samples.
OOD samples can be easily obtained for the classification task but not as 
straightforward for object detection
due to sensor incompatibility between datasets and lack of labels.
It means point clouds from different LiDAR
sensors can differ significantly in terms of beam arrangement and
intensity values, making it difficult to directly utilize data from
different datasets.
Moreover,
LiDAR-based 3D object detection datasets only provide labels for a handful of foreground classes
relevant to autonomous driving (e.g., cars, pedestrians, and cyclists). This
makes it challenging to gather diverse OOD objects from existing datasets even though the LiDAR sensors are compatible. In order to tackle aforementioned challenges:

\begin{enumerate}
    \item We propose a definition of OOD for object detection.
    Our analysis of OOD detection for object detection identifies six types
    of OOD objects with respect to the
    FG classes. We focus on the three of them for which the detector
    produces detections.
    \item We adapt and extend six
    existing OOD detection methods from classification to object detection.
    We design a feature extraction method for individual objects
    for methods that require a feature map as input.
    \item We propose a simple yet effective method to generate OOD objects
    to augment the existing dataset and extensively evaluate our 
    proposed OOD detection methods.
    \item We evaluate our OOD detection and object generation methods and demonstrate that different OOD detection methods have
biases toward detecting specific types of OOD objects. Thus, the best practices previously identified for OOD detection in
image classification may not hold in all cases.
\end{enumerate}

As far as we are aware, we are the first to explore OOD detection for
LiDAR-based 3D object detection and we wish to aid and stimulate future research on this exciting topic.

\section{Related work}
\label{sec:related}
\begin{figure*}[t]
    \centering
    \includegraphics[width=0.8\textwidth]{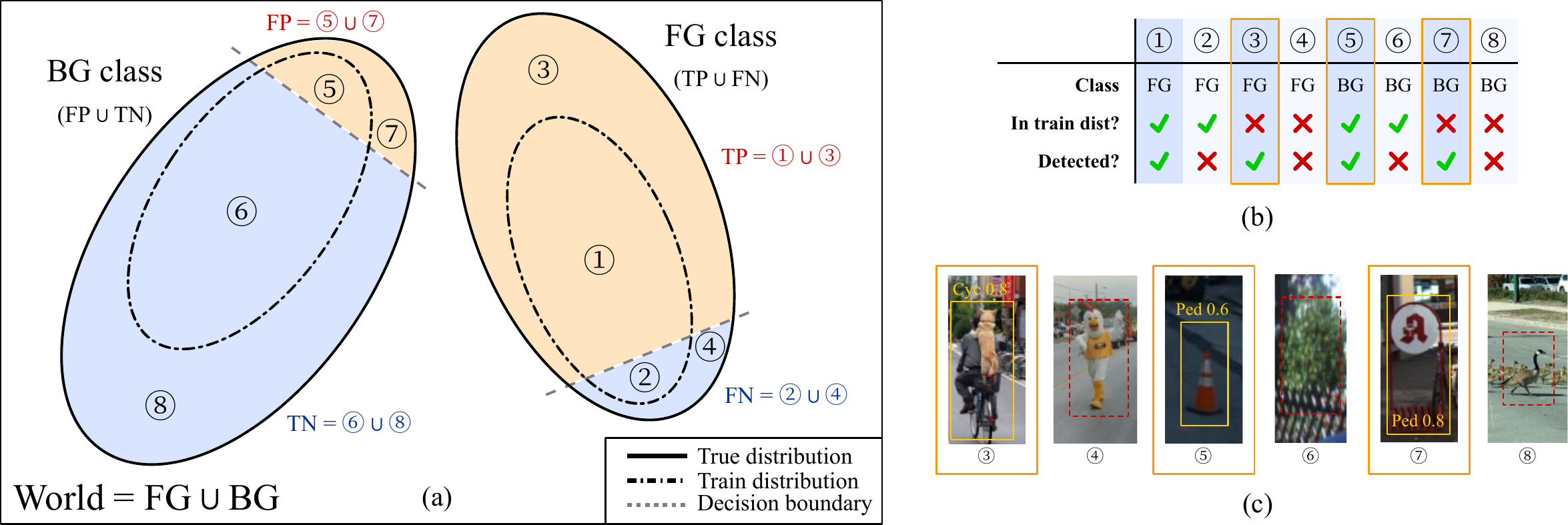}
    \caption{(a) Visualization of FG and BG object distributions and model decision
    boundary. (b) Classification of objects in object detection. (c) Examples of different types of OOD objects with respect to FG training distribution.}
    \label{fig:ood-diagram}
    \vspace{-0.2in}
\end{figure*}
\subsection{LiDAR-based 3D Object Detection} 
The goal of LiDAR-based 3D object detection is to produce 3D bounding boxes for the objects using LiDAR point cloud input.
Many architectures have been proposed to detect 3D objects based on raw point cloud, voxelized point cloud, or both
\cite{zhou2018voxelnet,lang2019pointpillars,shi2019pointrcnn,shi2020pvrcnn}.

In this work, we adopt PointPillars
\cite{lang2019pointpillars} as our object detector.
PointPillars first voxelizes point cloud input into
vertical pillars and extracts pillar features using
PointNet~\cite{qi2017pointnet}.
The pillar features are then processed by
a 2D convolutional backbone, followed by classification
and regression heads to generate the final predictions. 
We choose PointPillars due to its good performance and low
computational cost, which are essential for resource-constrained and
safety-critical systems.



\subsection{Out-of-Distribution Detection}

OOD detection has been investigated extensively under terms such as anomaly
detection, novelty detection or open-set classificaiton \cite{salehi2021unified-review-ood}. 
The majority of existing work focuses on classification
\cite{liang2017odin-cls,hendrycks2019-selfsupervised,rezende2015flow-cls,lee2018mahalanobis-cls,hendrycks2016baseline-cls,ren2019likelihoodood-generative-cls,zhang2020hybrid-cls,lee2017confidence-gan-cls,vyas2018ood-uncertainty-cls,laksh2016deepensemble-uncertainty-cls,li2017dropout-uncertainty-cls,hsu2020gen-odin-cls},
with some recent work on image segmentation
\cite{bevandic2018segmentation,angus2019segmentation,blum2019fishyscapes,marchal2020segmentation,williams2021segmentation}.

\citet{hendrycks2016baseline-cls} propose a baseline method for detecting OOD
inputs for deep image classifiers using max-softmax scores.
ODIN \cite{liang2017odin-cls} further improves the
baseline by applying temperature scaling and input perturbation.
\citet{hsu2020gen-odin-cls} propose Generalized-ODIN, removing the need for OOD
data for tuning. \citet{lee2017confidence-gan-cls} propose to train jointly a
classifier and a GAN network that generates OOD samples for training the
classifier. Further, \citet{lee2018mahalanobis-cls} use Mahalanobis distance of
the input sample to the nearest class-conditional Gaussian distribution
estimated from the in-distribution data as the sample's OOD score.



One-class classifiers such as OC-SVM \cite{scholkopf2001oc-svm} 
have been successfully applied to OOD detection as well.
Recently, \citet{abdelzad2019earlylayer-cls} propose to use an OC-SVM
trained with features extracted from an optimal layer to detect OOD samples. It
shows that features from earlier layers work well for OOD detection in image
classification. \citet{bishop1994novelty} suggests that a natural way to detect
OOD samples is to estimate the density of ID samples and check if samples are in
a low density area. Some works
\cite{an2015vaeood-generative,choi2018waicood-generative-cls,hendrycks2018ood-generative-cls,zisselman2020ood-generative-cls,ren2019likelihoodood-generative-cls}
also use generative models
\cite{kingma2013VAE,rezende2015flow-cls,oord2016pixelcnn} to detect OOD samples. 


Several works in image segmentation
\cite{bevandic2018segmentation,angus2019segmentation,marchal2020segmentation,williams2021segmentation}
have adapted OOD detection methods from image classification for OOD detection
at pixel level. They aim to reject the OOD pixels and 
improve segmentation performance.

In the automotive domain, \citet{blum2019fishyscapes} create a dataset to
benchmark OOD detection for image segmentation. \citet{nitsch2020out} evaluates
OOD detection for image detection applied to image patches. The OOD detector is
trained on KITTI and Nuscenes and tested in ImageNet.
\citet{wong2020identifying} propose a method to tackle open set semantic
segmentation in the 3D point cloud. The method can recognize and segment known
and unknown classes in 3D point clouds.

\section{Out-of-Distribution in Object Detection}
\label{sec:ood}
In this section, we propose a classification for ID and OOD 
objects in the context of object detection and provide examples for each case.

\subsection{Assumptions}

\textit{True Distributions:}
In object detection, all objects can be categorized as either foreground (FG)
or background (BG) objects. FG classes are what the object detector is trained
to detect, and BG classes are what the object detector is trained to ignore.
We assume the existence of true distributions for FG and BG objects
as indicated by the solid-line ovals in \Cref{fig:ood-diagram}a.
For the sake of simplicity, but without loss of generality,
we only consider a single FG class and a single BG class.

\textit{Training Distributions:}
FG or BG objects can be underrepresented
in the training dataset
due to the size of the dataset and how the data is collected. For instance,
a dataset collected exclusively in Europe would not contain objects
that are not sold in Europe. It results in a different distribution which
we refer to as the training distribution (dashed ovals in \Cref{fig:ood-diagram}a).

\textit{Decision Boundary:}
Given a training dataset, the object detector learns a decision boundary
that separates the FG and BG objects (dashed lines in \Cref{fig:ood-diagram}a). 
We assume this decision boundary is not
perfect due to the inherent ambiguity between FG and BG objects
(i.e., aleatoric uncertainty) and the shift between training
distributions and true distributions (i.e., epistemic uncertainty).
The decision boundary thus intersects the distributions which
results in four types of detections: true positive (TP), false positive (FP),
false negative (FN), and true negative (TN).

\subsection{Types of ID and OOD Objects:}
As shown in \Cref{fig:ood-diagram}a and \Cref{fig:ood-diagram}b, we can divide all objects into 
eight categories based on 1) if they are FG or BG, 2) if they are represented
by the training distribution, and 3) if they are detected by the object
detector. We show examples for the OOD categories in \cref{fig:ood-diagram}c. The categories are as follows.

\begin{enumerate}[label=\textcircled{\footnotesize{\arabic*}}]
\item Detected FG objects in the training distribution.
\item Missed FG objects in the training distribution.
\item Detected FG objects out of the training distribution. E.g., an unusual cyclist with a dog on his back correctly detected by the detector.
\item Missed FG objects out of the training distribution. E.g., a person wearing costume not detected by the detector.
\item BG objects (in training distribution) misdetected as FG.
E.g., a common traffic cone detected as a pedestrian.
\item Undetected BG objects (in training distribution).
E.g., a common bush not detected as any FG object.
\item BG objects (out of training distribution) misdetected as FG.
E.g., an uncommon bike rack detected as a person.
\item Undetected BG objects (out of training distribution).
E.g., an uncommon goose not detected as any FG object.
\end{enumerate}

In object detection, the goal is to detect FG objects, and BG objects are not localized. Thus, we define OOD objects for the FG training distribution, which are  \textcircled{\footnotesize3}-\textcircled{\footnotesize8}. 
Objects in \textcircled{\footnotesize3}-\textcircled{\footnotesize4}
include underrepresented or unseen FG objects,
and \textcircled{\footnotesize5}-\textcircled{\footnotesize8}
represent all BG objects that should not be detected by the object detector.

In this work, we only focus on detecting objects of type
\textcircled{\footnotesize3}, \textcircled{\footnotesize5}, 
and \textcircled{\footnotesize7}, and do not consider objects of one
FG class as OOD for another FG class.
Detecting \textcircled{\footnotesize4} is covered in the field of FN
detection~\cite{rahman-sign,introspective-fn} and is out of the scope of
this paper. Type \textcircled{\footnotesize6} and \textcircled{\footnotesize8}
are undetected BG objects, and thus do not affect the performance of
the object detector and other downstream tasks. We leave detecting these
types of OOD objects as future work as it would still be
beneficial to detect these cases for data collection and other purposes.

\section{OOD Point Cloud Generation} \label{sec:ood-dataset-gen} 
In this section, we describe our proposed method for generating point clouds with OOD objects. Our method inserts
OOD objects into existing point clouds,which can be used to evaluate OOD detection methods for LiDAR-based 3D object detection. 

We first build an OOD object database consisting of synthetic and real point clouds of individual OOD objects using LiDAR simulation and real sensor data from other compatible datasets.
The intensity values for the object point clouds are adjusted to
match the intensity distribution of the target dataset (i.e, in-distribution dataset).
As discussed previously, 
we focus on objects of type \textcircled{\footnotesize3},
\textcircled{\footnotesize5}, or \textcircled{\footnotesize7} 
defined in \Cref{sec:ood}.

To construct a point cloud with OOD objects,
we first randomly sample an OOD object point cloud from the
OOD object database and
a scene point cloud from the target dataset. Then, we insert the OOD object
into the scene via concatenation.
We do not simulate ``shadows'' caused by occlusions during insertion,
since the object
detector is already exposed to similar object insertions as part of the
augmentations during training. 

\begin{figure}[t]
\begin{center}
\includegraphics[width=0.8\linewidth]{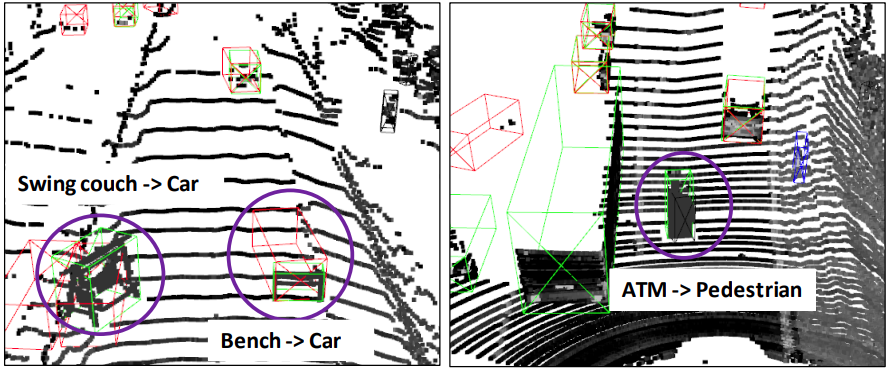}
\end{center}
\caption{
Examples of injected Carla objects detected as ID objects. Ground truth boxes are in green, Car predictions are in red, and Pedestrian predictions are in black.
}
\label{fig:detected_objects}
\vspace{-0.2in}
\end{figure}

During insertion, we preserve the OOD object's original distance to
sensor to keep the point cloud density consistent with its location, but we
randomly select the azimuth with respect to the sensor for more diversity.
We further make sure that
1) the inserted OOD object does not overlap with any
existing objects, 2) the inserted OOD object can be detected by the
object detector with a confidence score greater than some threshold $\tau$,
and 3) the predictions for the original objects in the scene are not
impacted by the new OOD object.
It is because low confidence predictions can be easily
filtered out by the object detector using its score threshold.
As a result, point clouds generated using this method
are tailored to a specific model. However, in practice
we observed that both synthetic and real OOD objects used in this work
can be consistently misdetected by multiple models.
\Cref{fig:detected_objects} shows some examples of the resulting point clouds. We can see that the object detector misdetects the inserted OOD objects as FG objects. The code for OOD dataset generation will be released after this paper is published.

Note that the dataset splitting method \cite{miller2021openset-uncertainty} for
generating an OOD dataset for object detection is not applicable in our context.
In contrast to image-based object detection datasets such as COCO
\cite{coco} and PASCAL \cite{pascal}, datasets for LiDAR-based 3D object
detection, such as KITTI \cite{geiger2013kittidataset},
Waymo \cite{waymo}, and NuScenes \cite{nuscenes}, contain only a handful of
labeled classes that are relevant to autonomous driving.
Furthermore, unknown objects in
\cite{miller2021openset-uncertainty} are not guaranteed to be detected
by the object detector.

\section{OOD Detection for 3D Object Detection}
We adapt five OOD detection methods from classification to LiDAR-based 3D
object detection, namely, max-softmax \cite{hendrycks2016baseline-cls},
uncertainty estimation (predictive entropy, aleatoric entropy and mutual information) \cite{li2017dropout-uncertainty-cls}, Mahalanobis
distance \cite{lee2018mahalanobis-cls}, OC-SVM \cite{scholkopf2001oc-svm}, and
normalizing flows \cite{dinh2016realnvp}. Among these methods,
Mahalanobis distance, OC-SVM, and
normalizing flows can be applied to intermediate features.
We propose a feature extraction method
for object detection and evaluate the quality of feature maps extracted from
various layers for OOD detection.

\subsection{Feature Extraction}
\label{ssec:feature-extraction}


Feature extraction for the object detection task is different from
classification since each input can produce an arbitrary number of predictions.
Thus, using the entire feature map may inadvertently include features from other
predictions or environmental objects. Instead, we identify a single feature
vector within each feature map for each prediction.

In this work, we use
PointPillars \cite{lang2019pointpillars} as our object detector.
In PointPillars, a set of predefined anchor boxes is
assigned to each pixel in the backbone feature map.
Labels and regression targets are assigned
to the anchor boxes based on the overlap with FG objects.
To train the OOD detection methods, we first identify the anchor boxes with positive
FG labels. We then extract feature vectors at their corresponding pixel locations 
in the feature map.
It is possible for one object to
correspond to multiple feature vectors during training, in which case we treat them as
independent training samples and do not perform any aggregation. During testing,
we identify the anchors from which the final predictions are generated and use their
corresponding feature vector in the feature map for OOD prediction.

In addition to the final backbone feature map, we also extract features from three
intermediate layers after each convolution block in the backbone, which we denote
\textit{conv2x}, \textit{conv4x} and \textit{conv8x}.
For feature
maps that are smaller than the backbone feature map where the anchor boxes are
defined, we apply nearest neighbors upsampling and use the aforementioned method
to extract feature vectors for training and testing.

\subsection{OOD Detection Methods}

\textbf{Max-softmax:} we follow previous work \cite{hendrycks2016baseline-cls}
and consider the maximum predicted class probability
as the baseline method for OOD detection. Lower max-softmax score indicates
the prediction is more likely to be an OOD object.

\textbf{Uncertainty estimates:} aleatoric and epistemic uncertainty estimates
can be used as the OOD score \cite{li2017dropout-uncertainty-cls}. Similar to
previous work by \citet{difeng2018towards}, we modify
the 3D object detector to obtain uncertainty estimates via MC-Dropout \cite{gal2016dropout}.

\textbf{Mahalanobis distance:} the Mahalanobis distance (in logit / feature space) measures the distance
between a sample and class-conditional Gaussian distributions estimated from in-distribution data.
It has been used to detect OOD samples in classification task \cite{lee2018mahalanobis-cls}.
In this work, we adopt this method to 3D object detection and experiment with logit layer and multiple feature layers.


\textbf{OC-SVM:} OC-SVM learns a decision boundary between in-distribution
data and the origin in Hilbert space that maxizes the distance between the
origin and the decision boundary \cite{scholkopf2001oc-svm}.
The signed distance from the sample to the decision boundary can be used for OOD detection.
Similar to Mahalanobis distance, OC-SVM can be applied to logit / feature space.

\textbf{Normalizing flows:} normalizing flows is a density
estimation method \cite{rezende2015flow-cls}. It aims to learn a series of
differentiable bijections that map complex distributions of observed data
to simple distributions of latent variables. Normalizing flows can output
log probabilities of the input sample. A well-trained normalizing flows model
will assign high log-likelihoods for ID samples and low
log-likelihoods for OOD samples. Therefore, predicted log-likelihoods can be
used as an OOD score. In this work, we use RealNVP \cite{dinh2016realnvp} in feature space as an efficient
method to learn the density of features associated with ID samples. 

\section{Evaluation} 
\subsection{Experimental Setup} 

\textbf{Dataset:} we use the KITTI dataset~\cite{geiger2013kittidataset}, which
has 7481 frames with annotated 3D bounding boxes. We split the
official dataset into training and validation splits (3712 and 3769 samples,
respectively)~\cite{lang2019pointpillars}. We use the training split for
training object detectors and extracting feature maps for the OOD detection
methods.

\textbf{Object detector:} we train a PointPillars model
for three foreground classes: \textit{Car,
Pedestrian}, and \textit{Cyclist}. To estimate classification uncertainty
using MC-Dropout \cite{gal2016dropout},
we add one dropout layer with dropout probability of 0.5 after
each deconvolution block. The classification head is also modified to output
softmax distribution instead of sigmoid scores. The performance of the modified
model is on par with vainilla PointPillars.



\textbf{OOD evaluation datasets:} we use the method described in
\Cref{sec:ood-dataset-gen} with detection threshold $\tau=0.3$
to generate OOD object datasets. 
We gather OOD objects from different sources to
insert into the KITTI dataset. For synthetic objects, we use the Carla
simulation \cite{dosovitskiy2017carla}. For real objects, we use the KITTI
ignored objects, the KITTI False Positive (FP) objects, and weird vehicle
objects from the Waymo dataset.
The KITTI FP objects are background
objects classified as Pedestrians (see \Cref{fig:motivation}). We manually label
and categorize them into five classes: \textit{potted plant, bike rack, low
traffic sign, sidewalk sign}, and \textit{thin sign}. We manually identify
objects among the Waymo vehicle class that are not FG in KITTI, including
\textit{motorcycle, scooter, digger and excavator}.
\Cref{fig:detected_objects} depicts examples of Carla objects inserted in
scenes and detected by the object detector.

\begin{figure}
    \centering
    \includegraphics[width=\linewidth]{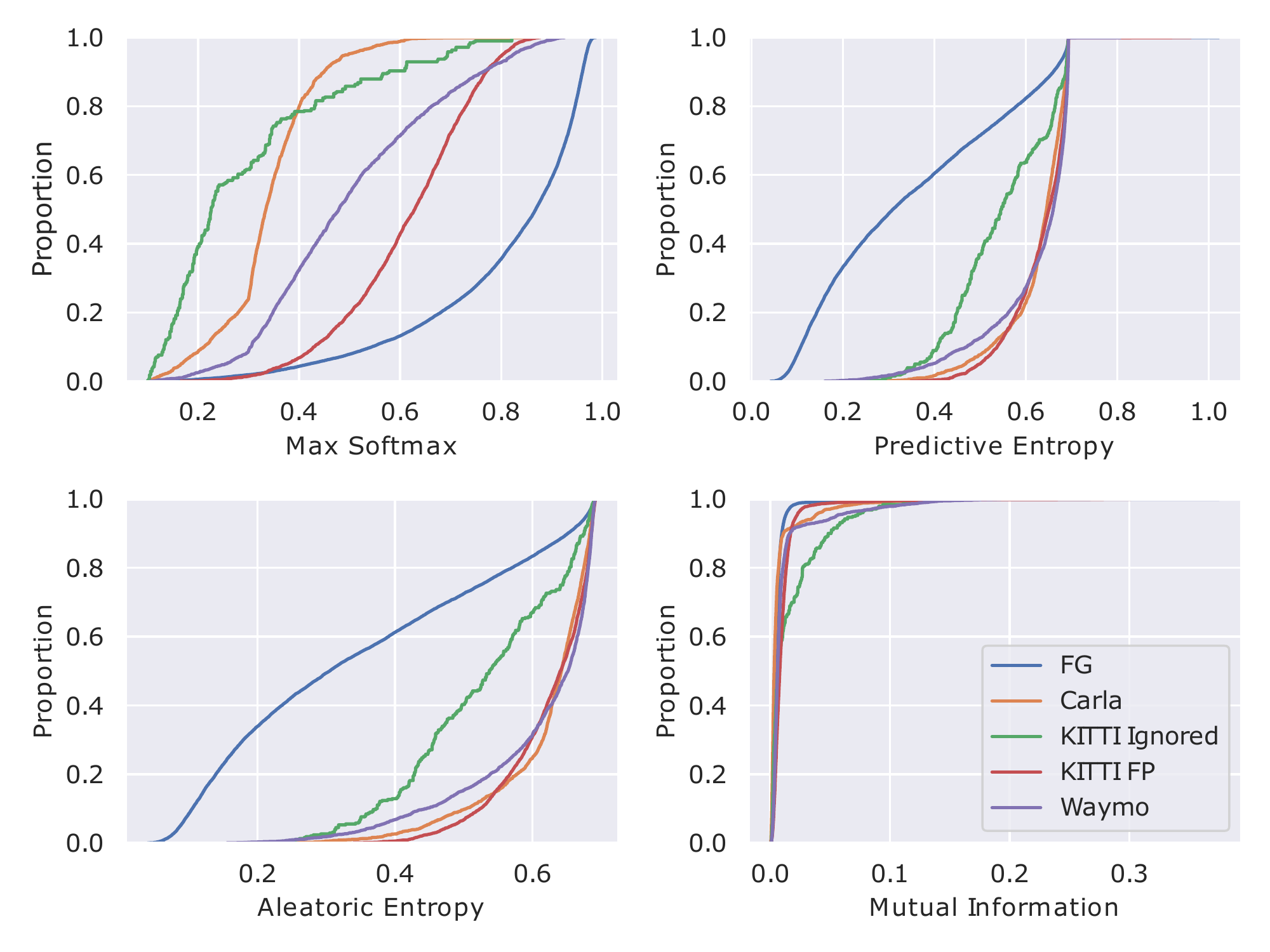}
    \caption{The CDFs of max-softmax scores, predictive uncertainty, aleatoric uncertainty and epistemic uncertainty of OOD object datasets.}
    \label{fig:uncertainty}
    \vspace{-0.2in}
\end{figure}

By collecting objects from different sources we ensure that our inserted objects
are diverse. \Cref{fig:uncertainty} shows the cumulative distribution of
softmax scores, predictive entropy, aleatoric entropy, and mutual information of
the inserted objects (i.e., OOD) and foreground objects (i.e., ID).
Our dataset covers objects with various ranges of uncertainties,
sofmax predicted scores and predictive entropy.



\textbf{OOD detection methods:} we
extract features from conv2x, con4x, con8x, and backbone to
train OC-SVM, Mahalanobis distance, and normalizing flows. For the OC-SVM
method, we train one OC-SVM with SGD per FG class and use the highest score as
the OOD score. We set $\nu = 0.01$, $\gamma = 2.0$ and train with a batch size
of 64 for 5 epochs. For Mahalanobis, we apply online mean/covariance update with
a batch size of 64 for 5 epochs. Normalizing flows is implemented using
RealNVP and trained with a batch size of 8 for 2320 steps.

\textbf{Evaluation metrics:} we adopt the evaluation metrics proposed by
Hendrycks et al. \cite{hendrycks2016baseline-cls}. We use AUROC and FPR at 95\%
TPR as our primary metrics due to the space limitation.




\begin{table*}[]
\centering
\scriptsize
\caption{The OOD detection results for all inserted OOD objects and each OOD object type. For each metric, we underline the best performing OOD method. For all results, we show the average and standard deviation over three sets of experiments.}
\label{tab:results_all}

\begin{tabular}{rc|ccccc|ccccc}
\toprule
\multirow{2}{*}{\textbf{OOD Method}} & \multirow{2}{*}{\textbf{Layer}} & \multicolumn{5}{c|}{\textbf{AUROC $\uparrow$}}                                                                   & \multicolumn{5}{c}{\textbf{FPR @ 95 TPR $\downarrow$}}                                                           \\
                                     &                                 & All                  & Carla                & Ignored              & KITTI FP                   & Waymo                & All                  & Carla                & Ignored              & KITTI FP                   & Waymo                \\ \hline
\textbf{Max Softmax}                 & \textbf{-}                      & 89.95\tiny$\pm$0.2                & {\ul \textbf{96.04\tiny$\pm$0.4}} & {\ul \textbf{95.58\tiny$\pm$0.5}} & 82.84\tiny$\pm$0.4                & 83.18\tiny$\pm$2.0                & 33.83\tiny$\pm$1.3                & {\ul \textbf{10.39\tiny$\pm$1.1}} & {\ul \textbf{21.23\tiny$\pm$3.6}} & 36.41\tiny$\pm$0.6                & 52.17\tiny$\pm$6.1                \\
\textbf{Predictive Entropy}          & \textbf{-}                      & 83.49\tiny$\pm$1.3                & 85.34\tiny$\pm$1.8                & 78.65\tiny$\pm$0.5                & 86.92\tiny$\pm$0.3                & 80.64\tiny$\pm$2.1                & 39.60\tiny$\pm$3.0                & 35.55\tiny$\pm$2.8                & 41.66\tiny$\pm$2.5                & 29.69\tiny$\pm$0.4                & 53.69\tiny$\pm$6.4                \\
\textbf{Aleatoric Entropy}           & \textbf{-}                      & 83.09\tiny$\pm$1.4                & 85.30\tiny$\pm$1.9                & 78.24\tiny$\pm$0.8                & 86.76\tiny$\pm$0.3                & 80.01\tiny$\pm$2.3                & 41.77\tiny$\pm$3.6                & 37.87\tiny$\pm$3.4                & 43.93\tiny$\pm$3.3                & 30.09\tiny$\pm$0.3                & 53.97\tiny$\pm$6.1                \\
\textbf{Mutual Information}          & \textbf{-}                      & 60.48\tiny$\pm$0.8                & 45.04\tiny$\pm$4.0                & 52.09\tiny$\pm$3.0                & 70.62\tiny$\pm$1.5                & 67.53\tiny$\pm$2.8                & 97.07\tiny$\pm$0.7                & 98.75\tiny$\pm$0.5                & 99.13\tiny$\pm$0.3                & 85.78\tiny$\pm$2.8                & 90.14\tiny$\pm$2.1                \\
\textbf{Mahalanobis}                 & \textbf{Backbone}               & 83.77\tiny$\pm$0.9                & 88.34\tiny$\pm$1.6                & 86.25\tiny$\pm$3.0                & 72.48\tiny$\pm$2.2                & {\ul \textbf{88.30\tiny$\pm$0.5}} & 51.01\tiny$\pm$1.0                & 37.49\tiny$\pm$2.2                & 54.35\tiny$\pm$11                 & 59.52\tiny$\pm$3.7                & 43.53\tiny$\pm$3.7                \\
\textbf{OC-SVM}                      & \textbf{Conv8x}                 & 65.44\tiny$\pm$3.0                & 66.49\tiny$\pm$2.9                & 66.95\tiny$\pm$2.9                & 63.22\tiny$\pm$3.0                & 64.74\tiny$\pm$2.3                & 66.20\tiny$\pm$5.0                & 63.49\tiny$\pm$5.6                & 62.58\tiny$\pm$5.7                & 72.03\tiny$\pm$6.8                & 67.81\tiny$\pm$3.7                \\
\textbf{Normalizing Flows}                     & \textbf{Backbone}               & {\ul \textbf{90.28\tiny$\pm$0.5}} & 93.63\tiny$\pm$0.5                & 85.81\tiny$\pm$1.0                & {\ul \textbf{91.71\tiny$\pm$1.0}} & 85.44\tiny$\pm$0.9                & {\ul \textbf{26.72\tiny$\pm$1.9}} & 19.90\tiny$\pm$0.7                & 31.17\tiny$\pm$0.5                & {\ul \textbf{16.25\tiny$\pm$1.3}} & {\ul \textbf{33.03\tiny$\pm$2.4}}\\
\bottomrule
\end{tabular}
\vspace{-0.2in}
\end{table*}

\subsection{Results and Discussion}
The results for all inserted OOD objects as well as each
OOD object type are shown in \Cref{tab:results_all}.
We apply
class-balanced resampling of ID and OOD samples
during the evaluation to remove any potential biases.
For Mahalanobis distance, OC-SVM, and normalizing flows, we report the results
of the layer with the best OOD performance.
In total, we
repeat dataset generation, model training, and evaluation process three times
and average the results.  

\textbf{Overall observations: }
normalizing flows with backbone features and max-softmax achieve the
best OOD detection performance. The performance of
OC-SVM and epistemic uncertainty (mutual information) via MC-Dropout are
significantly lower than other methods, suggesting that they may not be suitable
for OOD detection in 3D object detection context.

\textbf{Performance of different feature layers: }
the best performing layer for OOD detection can vary for different OOD
detection methods. In our experiments, the backbone features are optimal for
both Mahalanobis distance and normalizing flows, whereas OC-SVM works best with
conv8x features. 

Our results are different from existing work on image classification.
Using logits with Mahalanobis distance as proposed by
\citet{lee2018mahalanobis-cls} is out-performed by the backbone features by a
large margin in our experiments. Similarly, our OC-SVM model with early layer 
features could not achieve good separation between ID and OOD objects as observed by
\citet{abdelzad2019earlylayer-cls} in image classification.

This suggests that the optimal layer for each method can depend
on the specific model architecture or feature map distribution. Existing results
for image classification tasks may not transfer to other tasks with specialized
architectures such as PointPillars.

\textbf{Performance on different types of OOD objects: }
the performance of each OOD detection method can vary significantly for
different types of OOD objects. For instance, normalizing flows has 9\%
higher AUROC and 20\% lower FPR compared to max-softmax for KITTI FP objects
and has comparable results for Carla and Waymo
objects. However, it becomes
much worse than max-softmax for KITTI ignored objects with 10\% difference
in both AUROC and FPR. This can also be
observed for other OOD methods.

This variation can be due to the noticeable differences in uncertainty and
max-softmax probability between different OOD objects
(\Cref{fig:uncertainty}). The main observations are:  i) the max-softmax
scores and uncertainty estimations of the OOD objects can vary across
different OOD types, which affects the performance of these methods, ii)
Mahalanobis distance applied to the backbone layer is relatively stable
across different types of OOD. However, for KITTI FP objects with both high
max-softmax score and high uncertainty, its performance becomes
significantly worse, and iii) for normalizing flows with backbone layer, OOD
objects with higher aleatoric and lower epistemic uncertainty (Carla and
KITTI FP) have better performance.

In our experiments, we do not have a single OOD detection method that works
consistently well for all types of OOD objects.
This shows that depending on the type of OOD objects and their
characteristics, the optimal OOD detection method could be different. A
similar observation in the classification context is also noted by
\citet{kaur2021alloutlier}, who suggest that a combination of multiple OOD
detection methods may be needed to cover different types of OOD. We leave
in-depth investigation and evaluation of combined OOD detection methods for
future work.

\begin{figure}
    \centering
    \includegraphics[width=\linewidth]{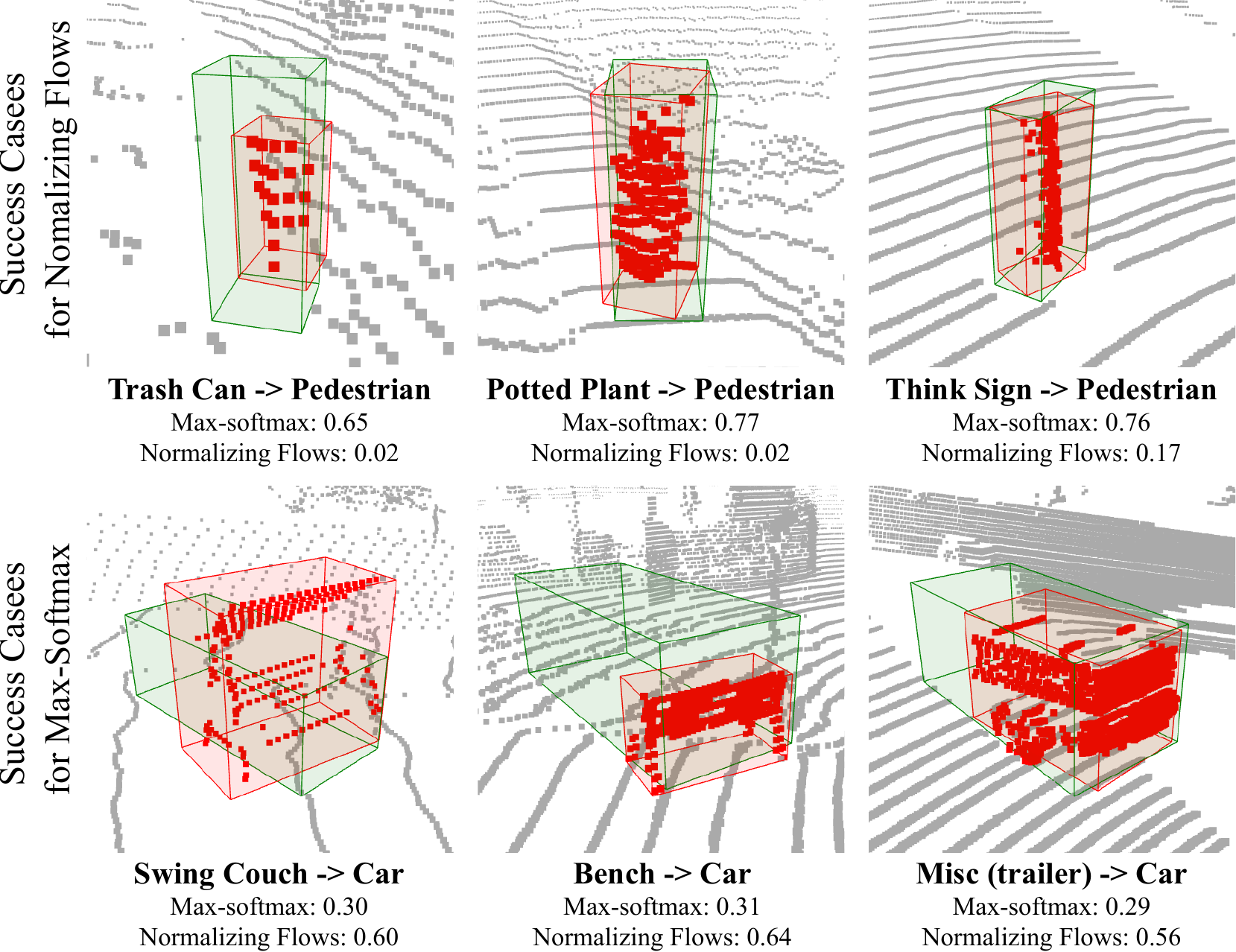}
    \caption{
    Qualitative results for max-softmax and normalizing flows. 
    Red boxes and points represent OOD objects, green boxes are predictions from the object detector.
   First row shows examples where normalizing flows successfully detects the OOD objects (indicated by low scores) but max-softmax fails (high max-softmax scores). Second row shows examples where normalizing flows fails and max-softmax detects the OOD objects successfully. }
    \label{fig:qualitative}
    \vspace{-0.2in}
\end{figure}
\textbf{Qualitative results: } we show in \Cref{fig:qualitative} success and failure OOD detections for
the two best-performing methods: max-softmax and normalizing flows.

\section{Conclusion and Future Work}
Deep LiDAR-based 3D object detectors play an important role in autonomous driving,
and making them robust to OOD objects is key for assuring the safety of such
systems. Although OOD detection has been defined and investigated extensively
for classification, it has not been explored for LiDAR-based 3D object
detection. In this paper, we define different types of OOD samples for object
detection and adapt the state-of-the-art OOD detection methods from image
classification to LiDAR-based 3D object detection. In order to use OOD detection
methods that rely on intermediate layers, we also propose a method for
extracting feature embeddings for the detected objects. To enable the evaluation
of the OOD detection methods, we propose a simple yet effective method to
generate OOD objects for LiDAR-based 3D object detectors. We evaluate the OOD
detection methods on the KITTI dataset augmented with a diverse set of real and
synthetic OOD objects, revealing a nuanced landscape of how the current OOD
detection methods perform in the context of LiDAR-based 3D object detection. The
results demonstrate that each method is biased toward detecting certain types of
OOD objects. Furthermore, the best practices proposed for image classification,
such as selecting features of specific layers for OOD detection, may not
transfer to object detection. We hope that our OOD dataset generation and
evaluation results will stimulate further research into effective OOD detection
for LiDAR-based 3D object detection. 


As future work, we plan to address the limitations of our proposed OOD dataset generation method. More specifically, we would like to 1) identify and insert unusual foreground objects (type \textcircled{\small 3} OOD), 2) ensure that the LiDAR intensity of the inserted objects is realistic and matches the intensity of the dataset, and 3) place OOD objects more realistically in the scene, possibly utilizing HD map information. 

Furthermore, we would like to extend our work by evaluating the OOD detection and object generation methods over other large-scale automotive datasets such as Waymo.
We are also interested in understanding the characteristics of OOD detection methods better and developing a combined OOD detection method that is not biased toward specific types of OOD objects.

{
    \small
    \bibliographystyle{IEEEtranN}
    \bibliography{macros,main}
}

\clearpage



\end{document}